\documentclass[runningheads]{llncs}

\usepackage[T1]{fontenc}
\usepackage{graphicx}
\usepackage{amsmath,amssymb}
\usepackage{booktabs}
\usepackage{xcolor}
\usepackage[numbers,sort&compress]{natbib}

\usepackage{tikz}
\usetikzlibrary{arrows.meta,positioning,calc,fit,backgrounds,shapes.geometric,decorations.pathreplacing}
\graphicspath{{figures/}}

\usepackage{hyperref}
\hypersetup{colorlinks=true, citecolor=blue, linkcolor=blue, urlcolor=blue}
\AtBeginDocument{}

\begin{document}
\frenchspacing

\title{Picking the Right Image to Classify: \\Reliable-Input Selection in Teledermatology}
\titlerunning{Picking the Right Image to Classify}


\author{
{Fabian Gröger}\inst{*, 1, 2} \and
{Marco Weishaupt}\inst{*, 2} \and
{Philippe Gottfrois}\inst{1, 3} \and
{Simone Lionetti}\inst{2} \and
{Linda Wermelinger}\inst{1, 2} \and
{Nipun Ranasekara}\inst{1, 2} \and
{Ludovic Amruthalingam}\inst{2} \and
{Alexander A. Navarini}\inst{\dagger, 1, 3} \and
{Marc Pouly}\inst{\dagger, 2}
}
\authorrunning{F. Gröger et al.}
\institute{
University of Basel \and
Lucerne University of Applied Sciences and Arts \and
University Hospital Basel\\
$^*$ equal contribution $\quad$ $^\dagger$ equal advising
}

\maketitle

\begin{abstract}
Dermatology models face distribution shifts in teledermatology settings, where submitted images differ from the training data in lighting, angle, distance, focus, and framing.
These test-time images are ordinary clinical photographs, but some fall outside the model's training conditions, leading the model to often misclassify them due to shifts in acquisition between training and deployment.
When multiple images of the same case exist (several photos of one patient or lesion), a natural way to improve accuracy is therefore to select the image the model is most likely to classify correctly.
We call this task \emph{reliable-input selection}.
An oracle that, for each case, selects a correctly classified image when one exists raises weighted F1 by about $20$ percentage points on average across six dermatology datasets and nine frozen backbones. This oracle is an upper bound that sees the labels, whereas a selector must choose blindly.
Capturing this gain in practice is hard. A selector that needs no pretraining data applies to any frozen model, including those whose data is not public. It must judge reliability from quantities the model exposes at inference: its embeddings, their norms, and its confidence.
We benchmark four such training-data-free selectors: the embedding norm, the neighborhood consensus among a case's images, the stability of the prediction under small perturbations, and the model's own confidence.
No training-data-free selector substantially narrows this oracle gap. The best of them is the model's own confidence, but it recovers only a small part of the gap on the clinical datasets.
A small labeled reference set does not help either: the best selector overall, a fusion of confidence and Mahalanobis distance, still leaves most of the gap.
To our knowledge, this is the first study to introduce and benchmark reliable input selection, a clinically important, unsolved task.

\keywords{Dermatology \and Reliable-input selection \and Acquisition shift.}
\end{abstract}

\section{Introduction}
\label{sec:intro}

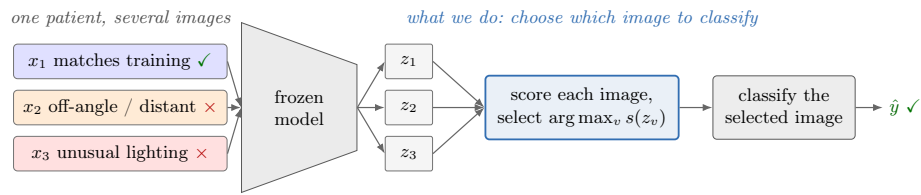
\begin{figure}[t]
\centering
\definecolor{barOurs}{RGB}{50,110,178}
\resizebox{\textwidth}{!}{%
\begin{tikzpicture}[
  font=\footnotesize, >={Latex[length=1.7mm]},
  img/.style={rounded corners=2pt, draw=black!55, fill=#1, minimum width=3.7cm, minimum height=0.6cm, align=center, inner sep=2pt},
  emb/.style={draw=black!55, fill=black!4, rounded corners=1pt, minimum width=0.82cm, minimum height=0.6cm, align=center},
  proc/.style={rounded corners=2pt, draw=black!65, fill=black!7, align=center, minimum height=1.1cm, inner sep=6pt},
  sel/.style={rounded corners=2pt, draw=barOurs, fill=barOurs!10, align=center, minimum height=1.1cm, inner sep=6pt, line width=0.9pt},
  ar/.style={->, black!60, line width=0.6pt},
]
\node[font=\footnotesize\itshape,text=black!70] at (0,3.55) {one patient, several images};
\node[font=\footnotesize\itshape,text=barOurs] at (8.0,3.55) {what we do: choose which image to classify};
\node[img=blue!12]   (i1) at (0,2.80) {$x_1$ matches training~\textcolor{green!45!black}{\checkmark}};
\node[img=orange!16] (i2) at (0,2.00) {$x_2$ off-angle / distant~\textcolor{red!70!black}{$\times$}};
\node[img=red!12]    (i3) at (0,1.20) {$x_3$ unusual lighting~\textcolor{red!70!black}{$\times$}};
\node[trapezium, trapezium left angle=70, trapezium right angle=70, shape border rotate=270,
      draw=black!65, fill=black!8, align=center, minimum width=0.5cm, minimum height=2.0cm, inner sep=2pt] (fm) at (3.1,2.00) {frozen\\model};
\foreach \i in {i1,i2,i3}{\draw[ar] (\i.east) -- (fm.west);}
\node[emb] (z1) at (5.0,2.80) {$z_1$};
\node[emb] (z2) at (5.0,2.00) {$z_2$};
\node[emb] (z3) at (5.0,1.20) {$z_3$};
\foreach \z in {z1,z2,z3}{\draw[ar] (fm.east) -- (\z.west);}
\node[sel] (sel) at (8.0,2.00) {score each image,\\select $\arg\max_v s(z_v)$};
\foreach \z in {z1,z2,z3}{\draw[ar] (\z.east) -- (sel.west);}
\node[proc] (cls) at (11.5,2.00) {classify the\\selected image};
\draw[ar] (sel.east) -- (cls.west);
\node[align=center,text=green!40!black] (out) at (13.6,2.00) {$\hat y$~\textcolor{green!45!black}{\checkmark}};
\draw[ar] (cls.east) -- (out.west);
\end{tikzpicture}}
\caption{\textbf{Reliable-input selection.} In teledermatology a patient submits several images of their skin condition, varying in body site and acquisition. Here one matches the model's training conditions (standardized) and two differ in angle, distance, or lighting (off-condition). A frozen model embeds each image as $z_v$ and classifies the standardized image correctly (\textcolor{green!45!black}{\checkmark}) but the off-condition images incorrectly (\textcolor{red!70!black}{$\times$}), silent failures that selection avoids. The task is to score the images and classify the selected one.}
\label{fig:overview}
\end{figure}

Dermatology models are often trained on controlled, standardized clinical photographs and applied afterwards to images captured at test time under different angles, distances, or lighting conditions.
Such acquisition shifts are a well-documented cause of accuracy loss~\cite{geirhos2018generalisation,taori2020measuring}.
These images are ordinary clinical photographs, not anomalies, so an out-of-distribution detector would not flag them and the model classifies them as usual. But because it was not trained for these conditions, it often misclassifies them, apparently confident~\cite{hendrickx2024machine}.

When multiple images of the same case are available, they can arise in several ways, three of which we study: repeated captures of one lesion, several viewpoints or modalities of one lesion, or several photographs of one patient covering different body sites, as in teledermatology. Rather than classifying an arbitrary image, one can select the image the model is most likely to classify correctly, a task we call \emph{reliable-input selection} (Fig.~\ref{fig:overview}).
Here we show that an oracle that, for each case, selects a correctly classified image when one exists raises weighted F1 by about $20$ percentage points on average without a stronger model or further training, a gain that current systems rarely exploit.

A selector that needs no pretraining data applies to any frozen model, and works from the frozen embeddings alone, with at most a small set of labeled images to calibrate a probe.
We benchmark four such training-data-free selectors, together with reference-set selectors that may also use a small labeled set, across six datasets and nine backbones, and the results are largely negative.
No selector recovers much of the gap. The model's own confidence helps on the multi-image clinical datasets but does not beat a fixed best view, the simple strategy of always using the single strongest acquisition type, where such a type is identifiable. Even a small labeled reference set leaves most of the gap.

Our contributions are:
(1) To our knowledge, we are the first to introduce and benchmark \emph{reliable-input selection}: choosing, among a case's images, the one a deployed dermatology model classifies most reliably, without its pretraining data.
(2) We quantify a large oracle gap (about $20$ percentage points of weighted F1 on average over six datasets) and show that four training-data-free selectors leave most of it. The strongest, classifier confidence, recovers at most about a quarter of the gap and does not beat a fixed best view where one is available.
(3) We trace the difficulty to the task itself rather than to a specific backbone or a simple methodological fix, and even a small labeled reference set helps little: the best such selector, a fusion of confidence and a class-conditional Mahalanobis distance, also recovers at most about a quarter of the gap.

\section{Related Work}
\label{sec:related}

\textbf{Generalization under acquisition shift.}
A large body of literature documents that accuracy drops when test inputs deviate from the training distribution.
\citet{geirhos2018generalisation} showed that networks matching humans on clean images are far less robust to image distortions. \citet{taori2020measuring} showed that robustness to synthetic perturbations does not transfer to natural shifts and that the main known remedy is training on larger, more diverse data. These studies establish that shift degrades accuracy, but none select among several available images for a given case at inference time. To our knowledge, no prior work studies this choice, and we are the first to introduce and benchmark it.
Test-time adaptation~\cite{liang2025comprehensive} updates the model to fit each shifted input, and multi-modal priors~\cite{zhou2025robust} add side information. Both change the model or its training, whereas we leave the model fixed and only choose which of a case's images to classify.

\textbf{Confidence, OOD detection, and sample rejection.}
Selective prediction equips models with a reject option, abstaining on inputs deemed unreliable~\cite{geifman2017selective,hendrickx2024machine}, often via confidence or out-of-distribution scores such as maximum softmax probability~\cite{hendrycks2017baseline}, ODIN~\cite{liang2018odin}, and Mahalanobis distance~\cite{lee2018mahalanobis}. Our task differs in two ways. We do not abstain but select among several valid images of one case, so the case is always processed. And the failure is not an anomaly: an unfamiliar but valid acquisition produces an ordinary, in-distribution embedding that outlier and density scores miss, yet the model misclassifies it.

\textbf{Foundation models and video in dermatology.}
Domain-specific foundation models provide strong dermatology embeddings with little labeled data~\cite{yan2025multimodal,kiraly2024health,yan2025derm1m}, while audits of the field's data basis reveal wide coverage gaps across skin tones and conditions~\cite{groger2025global}, which makes deployment-time reliability important. Video pipelines exploit redundancy across frames for more resilient detection~\cite{ahmed2025advanced}, and selecting reliable frames is a direct companion to such systems.

\section{Methods}
\label{sec:methods}

\subsection{The Reliable-Input Selection Task}
\label{sec:data}

Each case has $V$ images $\{x_1,\dots,x_V\}$, where a case is a single lesion or, in the per-patient datasets below, one patient. A frozen backbone maps them to embeddings $\{z_1,\dots,z_V\}$, a selection rule picks one index $v$, and the prediction on $x_v$ is reported. The metric is the final downstream performance of the selected images, which we report as weighted F1 because the diagnostic classes are imbalanced.
We use six publicly available dermatology datasets, two in each of three multi-image regimes, and our primary focus is the per-patient regime, the closest match to real-world teledermatology. \emph{Multiple images per patient} (different body sites): \emph{PASSION}~\cite{gottfrois2024passion}, patient-submitted images from sub-Saharan Africa across Fitzpatrick skin types III--VI ($1{,}022$ cases, $4$ conditions, $2$--$18$ images per case, median $3$), and \emph{DermaCon-IN}~\cite{madarkar2026dermacon}, a clinical collection from India ($1{,}457$ cases, $8$ classes, $2$--$13$ images), where a case is a patient and selection is per-patient image triage rather than a choice among aligned views. \emph{Distinct viewpoints or modalities of one lesion}, the aligned datasets and the only ones for which a fixed best view is defined: \emph{SCIN}~\cite{ward2024crowdsourcing}, whose cases give three images of one lesion from distinct viewpoints (close-up, at an angle, at a distance), from which we take the highest-weighted expert label, keep the $20$ most frequent conditions, and drop records lacking all three views or a label, leaving $1{,}045$ lesions, and \emph{derm7pt}~\cite{kawahara2019sevenpoint}, a clinical and a dermoscopic image of each lesion. \emph{Repeated captures of one lesion}: \emph{HAM10000}~\cite{tschandl2018ham10000} (repeated dermoscopy, $1{,}956$ lesions, $2$--$6$ captures) and \emph{PAD-UFES-20}~\cite{pacheco2020padufes} (repeated smartphone photos, $512$ lesions, $2$--$8$ captures).

\subsection{Backbones}
\label{sec:backbones}

We evaluate nine backbones spanning multiple pretraining paradigms, with no fine-tuning. These include PanDerm~\cite{yan2025multimodal} (dermatology, masked image modeling), MONET~\cite{kim2024monet} (contrastive CLIP), supervised~\cite{dosovitskiy2021vit} and masked-autoencoding~\cite{he2022mae} ImageNet ViTs, as well as DINO~\cite{caron2021dino} and DINOv2~\cite{oquab2024dinov2} (self-distillation). For DINOv2, we test four variants: plain, register~\cite{darcet2024registers}, patch-mean, and register-plus-patch-mean.
Spanning these paradigms tests whether the difficulty is a property of one model or of the task. None of the backbones is trained on our evaluation datasets: the ImageNet backbones use no dermatology data, PanDerm is pretrained on a corpus that excludes public benchmarks~\cite{yan2025multimodal}, and MONET on dermatology image-text pairs drawn from the medical literature~\cite{kim2024monet}.

\subsection{Selectors}
\label{sec:scores}

A selector assigns each image a score and takes the per-case $\arg\max$. We group selectors by what they are allowed to use.

\textbf{Baseline and references.} Our baseline is \emph{random} (classify any of the case's images): the realistic default, and the only choice defined for every dataset. We also report two ways to use a case's images without selecting one: \emph{majority vote} (the most frequent prediction across the images, random tie-breaks) and \emph{soft vote} (the argmax of their mean predicted probability). Two non-deployable references bound what selection could achieve. \emph{Best fixed} always classifies the same acquisition type, the one with the highest average F1. It is defined only when every case shares the same set of acquisition types, such as SCIN's three viewpoints or derm7pt's two modalities, and is undefined when a case's images are not aligned this way. The per-case \emph{oracle} picks a correctly classified image when one exists, which marks the upper bound any selector could reach.

\textbf{Training-data-free.}
The realistic deployment setting, in which selectors use only the frozen embeddings and the given probe.
The \emph{embedding norm} selector scores each image by LevyScore~\cite{maes2025levyscore}, $\log\mathrm{pdf}\,\chi_K(\lVert z\rVert)$: the typicality of the norm $\lVert z\rVert$ under an isotropic Gaussian latent space, in which it follows a $\chi_K$ distribution that recent self-supervised objectives encourage~\cite{lejepa}.
\emph{Neighborhood consensus} scores each image by its mean cosine similarity to the case's other images and prefers the most central one, on the assumption that the more an image's embedding deviates from the rest, the less reliable its prediction.
\emph{Perturbation stability} prefers the image whose prediction changes least under small Gaussian perturbations of its embedding, on the assumption that a reliable prediction is locally robust.
\emph{Classifier confidence} is the probe's maximum softmax probability.

\textbf{Reference-set selectors (small labeled set).} To establish an upper reference for the benefit of a small labeled set, we also include two selectors fitted on the practitioner's own labeled set (still without the model's pretraining data). The \emph{Mahalanobis} selector favors the image that looks most typical of its predicted class, measured by the Mahalanobis distance from its embedding to the class mean and covariance. The \emph{fusion} selector sums this typicality score with the classifier's confidence, both standardized. The class-conditional mean and covariance are estimated on the same split that fits the probe (Sec.~\ref{sec:eval}), not a separate hold-out, so they use no labeled data beyond the probe's.

\subsection{Evaluation Protocol}
\label{sec:eval}

On the frozen features, we evaluate two standard readout probes, a linear (logistic) probe and a $k$NN probe ($k{=}5$, cosine distance), which is the established protocol for assessing frozen representations~\cite{caron2021dino,oquab2024dinov2}, and both yield the same trends here.
Per seed we draw a stratified $60$/$40$ train/test split of cases. Each probe is fit on the training split for one acquisition type and applied to every image of each held-out case.
For each backbone we average each metric over the $10$ seeds and training-view choices. Because the classes are imbalanced, we report weighted F1. Error bars show $\pm 1$ standard deviation across the nine backbones. Accuracy shows the same trends.

\begin{figure}[t]
\centering
\includegraphics[width=\textwidth]{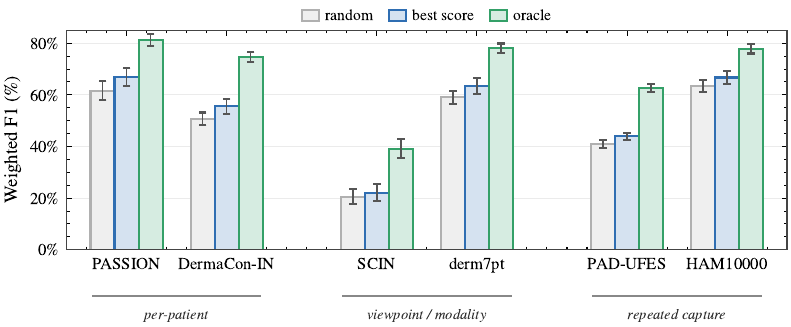}
\caption{\textbf{A large oracle gap that current selectors barely narrow.} Weighted F1 on the six datasets, grouped by multi-image regime (mean over nine frozen backbones and $10$ seeds, error bars show $\pm 1$ standard deviation across backbones). An oracle that picks a correct image per case (green) reaches F1 well above random selection (gray), whereas the best training-data-free selector (blue) recovers only a small part of that gap. We baseline against random throughout. A fixed best view is omitted, as it is defined only for the aligned datasets and is not deployable (see Sec.~\ref{sec:gap}).}
\label{fig:gap}
\end{figure}

\section{Experiments and Results}
\label{sec:experiments}

\subsection{The Oracle Gap}
\label{sec:gap}

We first measure the gain a perfect selection could recover. An oracle that, for each case, picks a correct image when one exists reaches F1 far above random selection (Fig.~\ref{fig:gap}), adding about $20$ percentage points on average. A fixed best acquisition, defined only on the aligned datasets (SCIN viewpoints, derm7pt modalities), sits just $2$ to $8$ F1 points above random. As it is not deployable, we use it as a realistic target rather than the baseline we measure gains against.

\subsection{Selectors}
\label{sec:battery}

We next ask how much of the gap any selector recovers, and find that none recovers much. Table~\ref{tab:battery} gives each method's weighted-F1 gain over random. The embedding norm adds essentially nothing and neighborhood consensus barely helps. The strongest training-data-free selector is the model's own confidence ($+3.7$ on average, up to $+5$ on the clinical datasets). A small labeled reference set helps a little more: a fusion of confidence with a class-conditional Mahalanobis score is the best selector ($+4.4$). Both recover at most about a quarter of the oracle gap, so reliable-input selection remains unsolved. Aggregating all of a case's images rather than selecting one leaves most of the gap too: soft voting reaches $+4.4$, as much as the best selector, and majority voting only $+1.6$ (it collapses to near-random where $V{=}2$ forces tie-breaks).

\begin{table}[t]
\caption{No selector recovers much of the oracle gap (weighted-F1 gain over \emph{random} selection, percentage points, mean over nine backbones, $10$ seeds). The strongest training-data-free selector is the model's own confidence. A fusion with a class-conditional Mahalanobis score (a small labeled reference set) is best overall, yet both leave the bulk of the oracle gap (last row) unexploited. \emph{Majority vote} and \emph{soft vote} aggregate a case's images instead of selecting one. Datasets are grouped by multi-image regime. The standard deviation across backbones is at most $3$ points.}
\label{tab:battery}
\centering
\resizebox{\textwidth}{!}{%
\begin{tabular}{lccccccc}
\toprule
 & \multicolumn{2}{c}{\emph{per-patient}} & \multicolumn{2}{c}{\emph{viewpoint / modality}} & \multicolumn{2}{c}{\emph{repeated capture}} & \\
\cmidrule(lr){2-3}\cmidrule(lr){4-5}\cmidrule(lr){6-7}
Method & PASSION & DermaCon-IN & SCIN & derm7pt & PAD-UFES & HAM10000 & mean \\
\midrule
majority vote           & $+3.9$ & $+2.6$ & $+1.3$ & $+0.2$ & $+0.7$ & $+1.0$ & $+1.6$ \\
soft vote               & $+6.3$ & $+6.0$ & $+2.4$ & $+4.8$ & $+3.6$ & $+3.6$ & $+4.4$ \\
\midrule
\multicolumn{8}{l}{\emph{training-data-free}} \\
embedding norm          & $+0.1$ & $+0.4$ & $-0.2$ & $+0.6$ & $+0.4$ & $+0.2$ & $+0.3$ \\
neighborhood consensus  & $+1.1$ & $+0.7$ & $+1.0$ & $+1.0$ & $+0.5$ & $+0.5$ & $+0.8$ \\
perturbation stability  & $+3.5$ & $+3.3$ & $+1.1$ & $+3.0$ & $+2.1$ & $+2.0$ & $+2.5$ \\
classifier confidence   & $+5.2$ & $+4.8$ & $+1.6$ & $+4.5$ & $+2.8$ & $+3.3$ & $+3.7$ \\
\midrule
\multicolumn{8}{l}{\emph{with reference set}} \\
Mahalanobis             & $+1.8$ & $+1.4$ & $+1.6$ & $+7.1$ & $+2.4$ & $+0.4$ & $+2.4$ \\
fusion (conf.\ $+$ geom.) & $+4.8$ & $+4.4$ & $+2.3$ & $+7.7$ & $+4.0$ & $+2.9$ & $+4.4$ \\
\midrule
\emph{oracle}           & $+19.8$ & $+24.0$ & $+18.7$ & $+19.0$ & $+21.7$ & $+14.5$ & $+19.6$ \\
\bottomrule
\end{tabular}}
\end{table}

\begin{figure}[t]
\centering
\includegraphics[width=\textwidth]{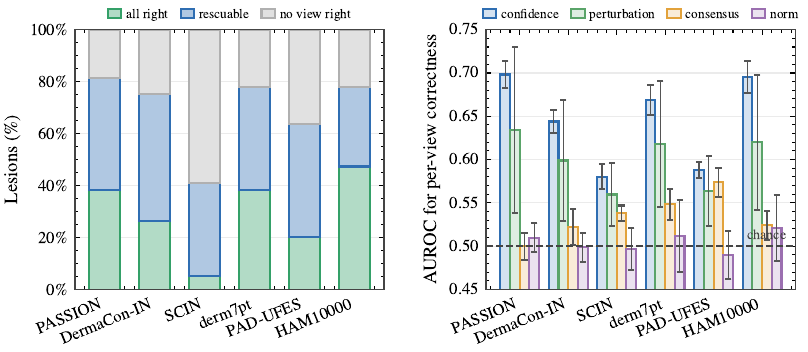}
\caption{\textbf{Why selection is hard: ample room, weak signal.} Left: every case is all-correct (selection irrelevant), no-image-correct (unrescuable), or mixed (rescuable). The mixed group is large, so there is room to rescue. Right: AUROC of each training-data-free score for per-image correctness on the mixed cases. The embedding norm and neighborhood consensus sit at chance. Only the classifier's confidence is clearly above it, and even then only moderately, too weak to exploit that room. All panels use all six datasets, in the same order as Fig.~\ref{fig:gap}. The dermoscopy datasets (derm7pt, HAM10000) show the same pattern as the clinical ones. Means over nine backbones and $10$ seeds. Error bars on the right panel show $\pm 1$ standard deviation across backbones.}
\label{fig:difficulty}
\end{figure}

\subsection{Sources of Difficulty}
\label{sec:why}

Selection can help only in the \emph{mixed} cases shown in Fig.~\ref{fig:difficulty} (left), where a case has both correctly and incorrectly classified images. The panel splits every case into all-correct (selection irrelevant), no-image-correct (no selector can help), or mixed.
This room is not the bottleneck: the mixed group is substantial across datasets, from $31\%$ (HAM10000) to $49\%$ (DermaCon-IN) of cases, so there is ample opportunity to rescue. (On SCIN a further $59\%$ of cases have no correct image at all and are beyond any selector's reach.)

The bottleneck is signal: whether any training-data-free score can identify the correct image among the mixed cases, which we measure as the AUROC of each score for per-image correctness (Fig.~\ref{fig:difficulty}, right).
The embedding norm sits at chance and neighborhood consensus barely above it. Only the model's own confidence is clearly above chance (AUROC $0.58$ to $0.70$, highest on PASSION), with perturbation stability, which derives from it, in between.
A usable per-image reliability signal exists, but lives only in the classifier's confidence, is at best moderate, and is too weak to exploit the room that exists.

This explains the small gains directly: a selector's accuracy is the all-correct fraction plus the mixed fraction times its hit rate on the mixed cases, and even the confidence selector raises that hit rate only modestly above random, far short of the oracle. Ample room with a weak signal leaves little to capture (Table~\ref{tab:battery}).

Selection is also not replaceable by a single fixed choice. On SCIN, the oracle's correct picks spread almost evenly across the three viewpoints ($36\%$, $33\%$, $31\%$). The best image changes from case to case, beyond any fixed viewpoint.

\subsection{Generality of the Result}
\label{sec:intrinsic}

The failure to recover the gap is tied to the dataset, not the backbone: the oracle gap's standard deviation across the nine backbones is under $2$ points, far smaller than its variation across datasets (Table~\ref{tab:battery}). It is also not removed by a simple methodological change.
It also recurs across all three regimes: the oracle gap is large in every one ($+14$ to $+24$ points), and the confidence selector recovers most in the per-patient regime ($+5$) and less elsewhere ($+3$).
The embedding norm is at chance across all nine backbones, because the images of a case are nearly indistinguishable in norm and position.
The gap is also not an artifact of keeping $20$ classes: restricting SCIN to the $K$ most frequent conditions ($K \in \{5, 10, 15, 20\}$) leaves the oracle $16$ (at the full $20$ conditions) to $23$ (at $5$) weighted-F1 points above the best training-data-free selector.

\section{Discussion and Conclusion}
\label{sec:discussion}

Reliable input selection is clinically relevant, since the choice of input alone adds about $20$ percentage points of weighted F1 on average across six datasets, led by the clinical teledermatology collections. Yet the task remains unsolved without the model's pretraining data: across nine backbones, no training-data-free selector recovers more than a small fraction of the gap, and even a small labeled reference set leaves most of it.

A selector can only succeed if the model exposes a reliable signal of its own competence, which current frozen encoders do not provide. A likely obstacle is the heavy augmentation used in self-supervised pretraining. While it pushes encoders toward invariance to the acquisition changes at issue here, the resulting embeddings remain far from truly invariant. Two images of the same case receive different predictions $34\%$ to $69\%$ of the time ($50\%$ on average). Their embeddings move, on average, about $0.62\times$ as far as embeddings of entirely different cases, indicating a within- to between-case cosine-distance ratio of $0.36$ to $1.0$ across datasets. A promising direction is therefore to train or adapt encoders whose embeddings track input reliability rather than discarding it. The absolute scores reflect benchmark difficulty (a hard $20$-class SCIN task, small usable subset, noisy labels), but our claim is relative: the oracle gap and the failure of every selector to recover it hold across datasets, backbones, and seeds.


\begin{credits}
\subsubsection{\ackname}
S.L., L.A., L.W., N.R., and M.P. are supported by the Swiss National Science Foundation (SNSF) under grant 20HW-1\_228541.

\subsubsection{\discintname} The authors have no competing interests to declare that are relevant to the content of this article.
\end{credits}

\bibliographystyle{splncs04nat}
\bibliography{references}

\end{document}